\documentclass[10pt,twocolumn,letterpaper]{article}

\usepackage{iccv}
\usepackage{times}
\usepackage{epsfig}
\usepackage{graphicx}
\usepackage{amsmath}
\usepackage{amssymb}

\usepackage[pagebackref=true,breaklinks=true,letterpaper=true,colorlinks,bookmarks=false]{hyperref}

\iccvfinalcopy 

\usepackage{comment}
\usepackage{multirow}
\usepackage{algorithm,algpseudocode}
\usepackage{url}

\usepackage{chngpage}
\usepackage{array}

\usepackage{bbding}
\usepackage{verbatim}
\usepackage{wasysym} 
\usepackage{xspace}
\usepackage{tabularx}
\usepackage[outercaption]{sidecap}    
\usepackage{dsfont}

\usepackage{booktabs}  

\usepackage{color, colortbl}
\definecolor{mygray}{gray}{0.95}
\definecolor{myred}{rgb}{1.0, 0.0, 0.0}
\definecolor{orange}{rgb}{1.0, 0.65, 0.0}

\usepackage{caption} 
\usepackage{amssymb}
\usepackage{pifont}
\makeatletter
\DeclareRobustCommand\onedot{\futurelet\@let@token\@onedot}
\def\@onedot{\ifx\@let@token.\else.\null\fi\xspace}

\def\eg{\emph{e.g}\onedot} 
\def\ie{\emph{i.e}\onedot}

\def\etal{\emph{et al}\onedot}
\DeclareMathOperator*{\argmax}{arg\,max}
\DeclareMathOperator*{\argmin}{arg\,min}
\DeclareMathOperator*{\st}{s.t.}
\makeatother

\makeatletter
\renewcommand{\paragraph}{
  \@startsection{paragraph}{4}%
  {\z@}{0.1ex \@plus 0.2ex \@minus .2ex}{-1em}%
  {\normalfont\normalsize\bfseries}%
}
\newcommand{\printfnsymbol}[1]{%
  \textsuperscript{\@fnsymbol{#1}}%
}
\makeatother

\def\iccvPaperID{7483} 

\ificcvfinal\fi

\begin{document}

\title{Self-Ordering Point Clouds}

\author{Pengwan Yang, Cees G. M. Snoek\printfnsymbol{1}, Yuki M. Asano\thanks {Equal last authors.}\\
University of Amsterdam\\
{\tt\small p.yang3@uva.nl}
}

\maketitle
\ificcvfinal\thispagestyle{empty}\fi

\begin{abstract}
   In this paper we address the task of finding representative subsets of points in a 3D point cloud by means of a point-wise ordering.
   Only a few works have tried to address this challenging vision problem, all with the help of hard to obtain point and cloud labels. Different from these works, we introduce the task of point-wise ordering in 3D point clouds through self-supervision, which we call self-ordering. We further contribute the first end-to-end trainable network that learns a point-wise ordering in a self-supervised fashion. It utilizes a novel differentiable point scoring-sorting strategy and it constructs an hierarchical contrastive scheme to obtain self-supervision signals. We extensively ablate the method and show its scalability and superior performance even compared to supervised ordering methods on multiple datasets and tasks including zero-shot ordering of point clouds from unseen categories.
\end{abstract}

\section{Introduction}

\begin{figure}[t!]
	\centering
	\includegraphics[width=1.0\columnwidth]{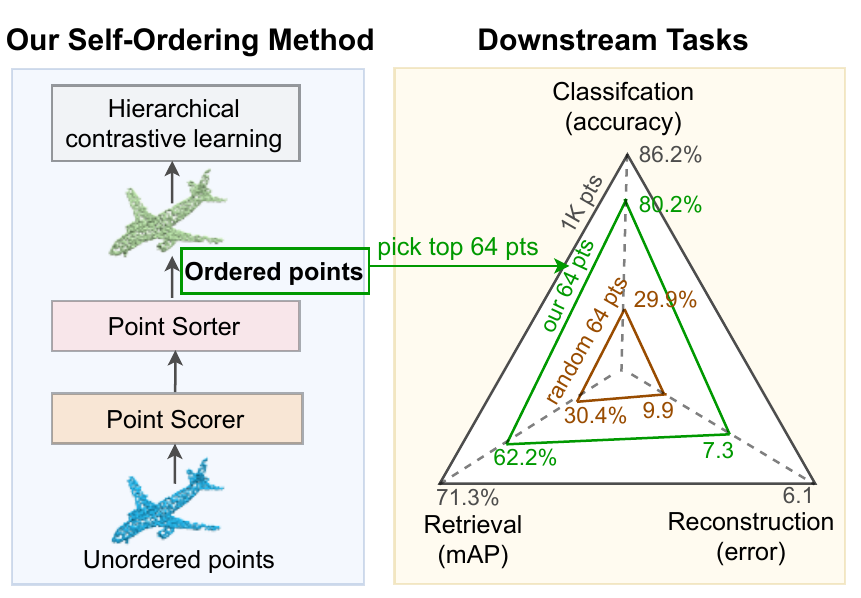}
	\caption{\textbf{Self-supervised learning to order point clouds.} We develop the first self-supervised method to order 3D points in terms of their importance. 
	By utilizing a novel Point Scorer and differentiable Point Sorter module, we leverage a hierarchical contrastive loss for training to output a point-wise ordering without the need for any labels. This can be used to select subsets that not only outperform previous heuristic methods but even the current, supervised, state of the art on downstream tasks. 
 }
	\label{fig:setting}
\end{figure}

The goal of this paper is to order the points in a 3D point cloud, so as to enable compute-, memory- and communication-efficient deployment. 
This is relevant for autonomous driving~\cite{cui2021deep,zeng2018rt3d}, scene understanding~\cite{behley2019semantickitti,nie2021rfd}, and virtual reality~\cite{yu2022method,el2019virtual}, to name just three of the many applications. The key problem of this modality of data is the sheer dimensionality of the input: while each point is usually only 3-dimensional, the number of points per scene or object can easily range from thousands to even millions~\cite{dai2017scannet,zamir2016generic}. As even the original data often contains highly redundant points, methods have been employed to reduce the number of points for particular tasks to a more manageable size.
These include simple heuristics such as picking random points~\cite{qi2017pointnet} or selecting the ones with large Euclidean distances~\cite{li2018pointcnn,qi2019deep,qi2017pointnet++}. 
More recently, methods have been proposed that order points based on their importance~\cite{zheng2019pointcloud,dovrat2019learning,lang2020samplenet}.
For example, Lang~\etal~\cite{lang2020samplenet} utilize a supervised network to optimize for smaller subsets of point clouds that can still yield good performance on various tasks. However, this method requires supervised annotations in the form of labels, which carry with it two main disadvantages. First, is the limited scalability -- as obtaining human-annotations, as well as developing a systematic annotation scheme -- is expensive.
Second, labelling data captured in-the-wild (such as from a vehicle-mounted LiDAR sensor) is highly ambiguous if there is more than one object present: given a point cloud with 10K points, finding which one belongs to the street, a car, or another car is both difficult and tedious. To avoid the label-dependency of existing work, we propose in this paper the first self-supervised point cloud sorter (see Figure~\ref{fig:setting}).

For self-supervised learning in the image domain, techniques such as clustering~\cite{caron2018deep, asano2019self} and contrastive learning~\cite{wu2018unsupervised, he2020momentum, chen2020simple} have recently shown great potential in learning generalizeable representations by designing suitable unsupervised loss functions. However, these methods are not directly applicable to the problem of point ordering due to two main differences. 
First, compared to the image domain, our goal is not representation learning \textit{per-se}, but rather to leverage an unsupervised training signal for finding a reduced set of points that are representative for many downstream applications.
Indeed, the difficulty lies in designing an effective module that can conduct the ordering -- the key trainable component in our method -- in a differentiable manner.
Finally, in particular for contrastive learning, the generation of positive pairs is of key importance. While in the image domain this is trivially accomplished with augmentations, for point clouds this is generally difficult~\cite{chen2020pointmixup} and further complicated for the important case of having a very low number of points. 

In this paper, we address these two challenges by ordering points in a 3D point cloud in a self-supervised manner -- a setting we call \textit{self-ordering}. The first challenge we address by front-loading a novel differentiable point scoring and sorting module that produces subsets, for which the self-supervised loss is then used for tuning.
The second challenge we solve by naturally setting different strict subsets of particular point clouds as positive pairs -- a choice that arises naturally and leads to a simple but novel hierarchical contrastive loss. As the first to tackle this setting in an unsupervised manner, we show that our method can not only produce good orderings of points, but even outperform supervised methods, when evaluated on a range of downstream tasks and datasets.

Overall, we make three contributions in this work.
\begin{itemize}
\itemsep0em 
    \item We propose the setting of self-ordering point clouds by self-supervised learning, in which each point in a 3D cloud  is ranked according to its importance without relying on any annotations. 
    \item We develop an end-to-end trainable approach to learn a point-wise ordering. For this, we develop a novel differentiable point scoring-sorting strategy and construct a hierarchical contrastive scheme to obtain self-supervision signals.
    \item We extensively ablate the method and show its scalability and superior performance even compared to supervised point ordering methods on multiple datasets and tasks including zero-shot ordering of point clouds. 
\end{itemize}

\section{Related work}

\paragraph{Self-supervised point cloud representations.} Most self-supervised learning works focus on learning single, global 3D object representation with applications to classic vision tasks,~\eg, classification~\cite{huang2021spatio,hassani2019unsupervised,sanghi2020info3d}, reconstruction~\cite{sauder2019self,hassani2019unsupervised,liu2022spu}, or part/semantic segmentation~\cite{huang2021spatio,eckart2021self,fu2022distillation}. Recent works start to demonstrate potential for more high-level tasks or complex scenes,~\eg, Mersch~\etal~\cite{mersch2022self} predicted future 3D point clouds given a sequence of point clouds. Zhang~\etal~\cite{zhang2021self} proposed a self-supervised method to build representations for scene level point clouds without the need for multiple-views of the scene. Wang~\etal~\cite{wang2021self} proposed a self-supervised schema to learn 4D spatio-temporal representations from dynamic point cloud clips by predicting their sequential orders without any ground-truth point-wise annotations. In this paper, our goal is not to learn a generic representation of point clouds but instead leverage self-supervision for the task of point cloud ordering.

\paragraph{Point cloud contrast construction.} Unlike conventional data structures such as images or videos, raw point clouds are unordered sets of vectors. This difference naturally poses challenges for designing contrastive learning methods on point clouds. 
There are three main ways of creating positive and negative pairs (contrasts) for learning: 1) use contiguous chunks of a 3D point cloud as positives~\cite{eckart2021self,liu2022spu}; 2) treat geometric transformation as a second view to generate positives~\cite{huang2021spatio}; 3) use natural or synthetic temporal sequences of point clouds~\cite{huang2021spatio,li2021self} as positives.
Most works used one or two of them,~\eg, Sanghi~\etal~\cite{sanghi2020info3d} use local chunks of 3D objects along with geometric transformation to build an effective contrastive learning. We do not use the common methods for contrast construction. 
Instead, we build a hierarchy of point clouds to generate a contrastive loss that operates on subsets inside the point cloud. These subsets in turn are not fixed or predefined, but are learned from our scorer network, and could thus be seen as a data-dependent view generator.

\paragraph{Differentiable top-k with optimal transport.} 
Several works~\cite{xie2020differentiable, cuturi2019differentiable, bonneel2016wasserstein,xie2020differentiable} studied the introduction of a regularization term in the optimization to make a selection operator differentiable from the perspective of optimal transport. 
Specifically, Cuturi~\etal~\cite{cuturi2019differentiable} propose an optimal transport problem for finding the k-quantile in a set of $n$ elements, 
while Xie~\etal~\cite{xie2020differentiable} parameterize the top-k operation as an optimal transport problem. 
Cordonnier~\etal~\cite{cordonnier2021differentiable} implement a differentiable top-k operator in the image domain to select the most relevant parts of the input to efficiently process high resolution images. Inspired by the above technical algorithmic contributions, we treat point sorting as an optimal transport problem. 
To bring these techniques to the 3D domain, we develop novel point scorer and sorter modules which are tailored to point clouds as inputs and allow for differentiable ranking of points in a cloud. Our novelty is not to improve the top-k selection theory but to apply it to a novel realistic problem using self-supervision.


\begin{figure*}[th]
	\centering
	\includegraphics[width=0.99\linewidth]{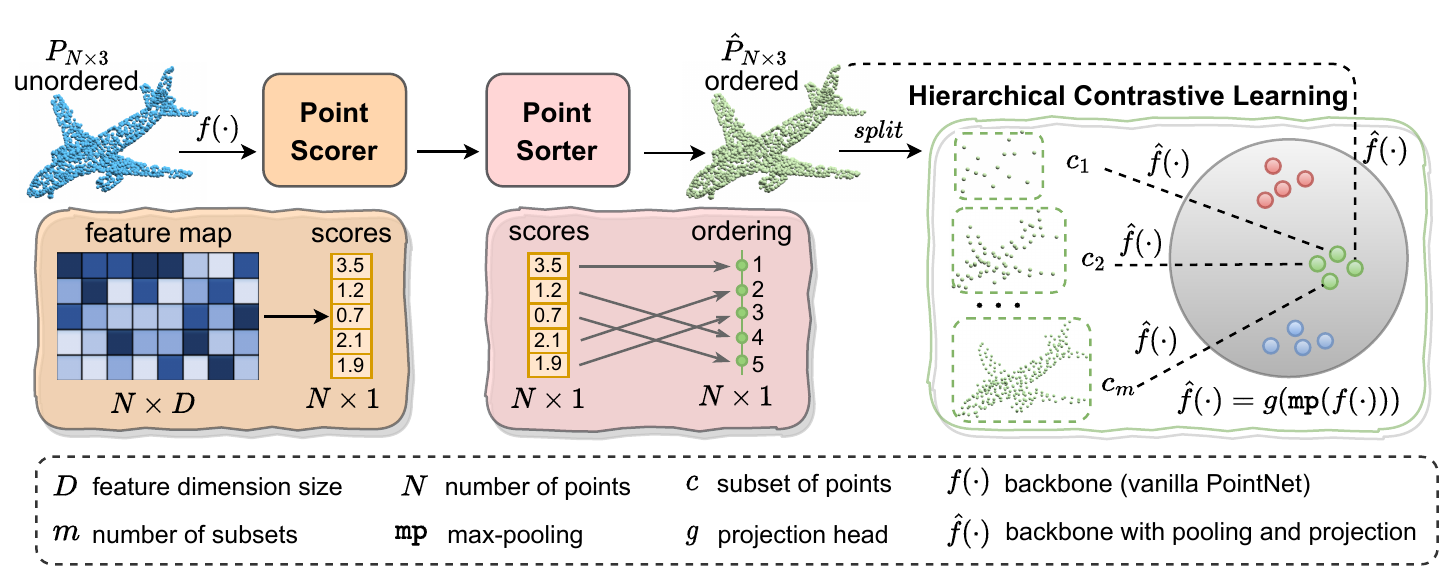}
	\caption{\textbf{Overview of the proposed model.} 
	The Point Scorer network predicts a score for each point in the 3D point cloud. The Point Sorter module transfers the scores of points into a ranking of points, and the Hierarchical Contrastive Learning module forces the ranking to be correct. Note that we order the point cloud without the need for any label supervision.} 
	\label{fig:architecture}
\end{figure*}

\paragraph{Point ordering with label supervision.}
Different from point cloud completion~\cite{wang2022softpool++} and reconstruction~\cite{achlioptas2017learning}, which change the point and cloud values, point ordering does not change the point and cloud values but reorders the points in a cloud.
So far, ordering points in a cloud has only been addressed using supervised data~\cite{zheng2019pointcloud,dovrat2019learning,lang2020samplenet}. 
Zheng~\etal~\cite{zheng2019pointcloud} assign each point a score by calculating the contribution of each point to the gradient of the loss function in a pre-trained model. Dovrat~\etal~\cite{dovrat2019learning} utilize a pre-trained task network as a teacher to optimize an ordered transformation of the original point cloud, then project the ordering information from the ordered transformation into the original point cloud. Lang~\etal~\cite{lang2020samplenet} follow Dovrat~\etal in utilizing a pre-trained task network but introduce a differentiable projection tool to obtain an even more precise ordering. We are inspired by both Dovrat~\etal and Lang~\etal, as we tackle the same task, but rather than using labels we order point clouds without any supervision.

\section{Method}

\paragraph{Problem formulation.}
We seek to order points in 3D clouds using self-supervision. 
We denote a point cloud as $P {=} \{p_i\}_{i=1}^{N}$ with $p_i \in \mathbb{R}^{3}$. 
Our goal is to find an ordering of the points $\gamma^{*} {=} ( i_{1},i_{2},\cdots,i_{N} ) $ solely using an unlabeled dataset of point clouds that minimizes the downstream objective function $\phi
$:
\begin{equation}
\gamma^{*} =\argmin_{\gamma}{\phi(S_n)},
\label{equ:problem_description}
\end{equation}
where various subsets of $S_n {=} \{p_{i_k}\}_{k=1}^{n},$ contain the top $n \,(\leq N)$ ranked points.

The challenge of this objective is two-fold.
First, the objective is defined in terms of a downstream objective function, yet the ordering $\gamma^*$ is generated solely from unlabeled data $P$. 
Note that this means our task is two levels of abstraction away from the downstream task: we neither have labels per point cloud, nor do we have point-level annotations.
Second, finding a right permutation is fundamentally a non-differentiable operation and directly learning this with gradient descent is difficult due to the discrete nature of permutations~\cite{zheng2019pointcloud,santa2017deeppermnet}.

In our method, we tackle both of these difficulties in turn by a) constructing a hierarchical self-supervised loss operating on subsets of 3D point clouds and b) constructing these via a differentiable scoring-sorting procedure.


\paragraph{Model overview.}
As illustrated in Figure~\ref{fig:architecture}, our model consists of a \textit{Point Scorer} network, a \textit{Point Sorter} module, and a \textit{Hierarchical Contrastive Learning} module. 
The Point Scorer outputs a floating point value for each point in the 3D point cloud, 
the Point Sorter module orders the points according to these scores in a differentiable manner, 
and the Hierarchical Contrastive Learning module provides the necessary gradients to ensure the ordering is meaningful.
The three modules are described in more detail below.

\paragraph{Point Scorer.} 
The scorer is designed for scoring the points in the cloud. It thus provides a mapping $f$ from a point cloud to a score vector, $f: P\rightarrow \mathbb{R}^{|P|}.$
More specifically, the original point cloud $P$ is firstly fed to a point cloud backbone $f(\cdot)$, yielding a  feature map $\mathcal{F} \in \mathbb{R}^{ |P| \times D}$, where $D$ is the feature dimension size. Next, the goal is to compute a score for each point from these features. For this, we denote $\mathcal{F}_i{=}\{f_{i1},f_{i2},\cdots,f_{iD}\}$ as the $D$-dimensional feature of the $i$-th point, and $f_{ij}$ as the $ij$-th element. 

To arrive at a score for a point $p$, we compute the contribution of a point to the global feature of the point cloud $P$. 
For a PointNet architecture~\cite{qi2017pointnet}, this global feature $\mathcal{G} = \{g_1, g_2, \cdots, g_D\}$  is computed by an order-invariant max pooling operation, 
\ie, 
\begin{equation}
    g_j = \mathop{max}\limits_{i=1,2,\cdots,N}f_{ij}, 
    \quad \mathrm{for } \,\,j\in \{1,2,\cdots,D\},
    \label{equ:g}
\end{equation}
or simply $\mathcal{G} {=} \texttt{mp}(\mathcal{F}, dim{=}0)$. 
Let $\mathrm{idx}_j$ denote the point which has the maximum value in the $j$-th dimension, \ie, 
\begin{equation}
    \mathrm{idx}_j=\argmax\limits_{i=1,2,\cdots,N}{f_{ij}},\quad \mathrm{for } \,\,j\in \{1,2,\cdots,D\},
    \label{equ:idx}
\end{equation}
The simplest metric to compute the importance of a point $i$ to the final feature would be to count how often its features ended up in the global feature:
\begin{equation}
    s_i=\frac{1} {D} \sum_{j=1}^{D}\delta({\mathrm{idx}_j,i}),
    \label{equ:s_i}
\end{equation}
where $s_i$ is the score of $i$-th point and $\delta({x,y})$ is a Kronecker-delta function, \ie, 
$\delta({x,y}) {=} 0 \text{ if } x {\neq} y$ and $1 \text{ if } x{=}y  $.
Intuitively, if one point's feature is descriptive of the whole global feature, it achieves a score of $1.0$.

To enable backpropagation, we replace the operations of Equations~\ref{equ:idx}-\ref{equ:s_i} with a differentiable approximation which we formulate as:
    
\begin{equation}
    s_i = \frac{1}{D}\sum_{j=1}^{D}2\cdot \sigma(f_{ij} - g_j),
    \label{equ:S_approx}
\end{equation}
where $\sigma$ is a sigmoid operation with temperature $\tau$,~\ie, $\sigma(x){=}\frac{1}{1+e^{-x/\tau}}$. By scaling with 2, the sigmoid outputs arrive at the interval [0, 1]. Again, it holds that if one point's feature $\mathcal{F}_i$ is equal to the global descriptor $\mathcal{G}$, $s_i{=}1.0$.
Note that from Equation~\ref{equ:S_approx}, $s_i\propto \Sigma_j(\sigma (f_{ij} -g_j)$, leading to the simple calculation of the score vector for all points as $\mathcal{S} {=} \texttt{sum}(\sigma(\mathcal{F}-\mathcal{G}), \texttt{dim}{=}1)$.

\paragraph{Point Sorter.}
The task of this module is to sort the points according to their scores in a differentiable manner.
For this, we apply a Top-K operator utilising optimal transport~\cite{xie2020differentiable,cuturi2019differentiable}. 
%
We first parameterize point sorting in terms of an optimal transport problem. 
We try to find a transport plan from discrete distribution $\mathcal{S}{=}[s_1,s_2,\cdots,s_N]^T$ to discrete distribution $\mathcal{B}{=}[1,2,\cdots,N]^T$. 
For this, we specify the marginals for both $\mathcal{S}$ and $\mathcal{B}$ as $\mu{=}\nu{=}\textbf{1}_N/N$ and
further denote $C\in \mathbb{R}^{N\times N}$ as the cost matrix with $C_{ij}$ being the cost of transporting mass from $s_i$ to $b_j$, which is the $j$-th element in $\mathcal{B}$. We take the cost to be the squared Euclidean distance,~\ie, $C_{ij}=(s_i-j)^2$.
Given this formulation, the optimal transport problem can be formulated as:
\begin{equation}
    \begin{split}
        \Gamma^{*}=\argmin\limits_{\Gamma\ge0}\left \langle C,\Gamma \right \rangle+ \epsilon h(\Gamma),\\ \st,\Gamma\textbf{1}_N = \mu, \Gamma^T\textbf{1}_N=\nu,
        \label{equ:transport}
    \end{split}
\end{equation}
where $\left \langle \cdot \right \rangle$ is the inner product, $h(\Gamma){=}\sum_{ij}\Gamma_{ij}\log\Gamma_{ij}$ is an entropy regularizer which can rule out the discontinuity and obtain a smoothed and differentiable approximation to the Top-K operator. 
Then the approximate optimal $\Gamma^{*}$ is referred to as the optimal transport plan. 
Finally, by scaling with $N$, we obtain \mbox{$\gamma^*{=}N\Gamma^*\cdot\mathcal{B}$} as the ordering of the points, where the point with the highest score is set to 1, the point of second biggest score to 2, and so on. 
This yields the sorted/ordered point cloud $\hat{P} \in \mathbb{R}^{N\times 3}$ for the original unordered point cloud $P$.

\paragraph{Hierarchical Contrastive Learning.} 
Now that we have a differentiable method for obtaining an ordered point cloud $\hat{P}$, the next step is to optimize this using self-supervision. We use a hierarchical scheme inside the ordered point cloud as a free supervision signal.

For this, we define a set of subsets with exponentially increasing cardinality, \ie, $\mathcal{C} {=} \{c_k\}_{k=1}^m$, with $|c_k| {=} \theta^{k}$, $m{=}\log_{\theta}(N)$ and $\forall k: c_k \subset c_{k+1}$. 
Here, $\theta$ governs the growth rate of the subset. 
The first subset contains the top $\theta$ points in $\hat{P}$, the second subset the top $\theta^2$ points and so on. 
We treat subsets $\mathcal{C} {=} \{c_k\}_{k=1}^m$ of $\hat{P}$ as positive pairs while negative pairs are constructed from subsets of other point clouds. 
Then we naturally arrive at the following multiple-instance NCE formulation~\cite{miech2020end}:
\begin{equation}
\mathcal{L}^{c_k}_{\mathrm{NCE}}(\hat{P})=-\log\frac{\sum_{c_k^+}\exp( \langle \hat{f}(\hat{P}), \hat{f}(c) \rangle /\varphi)}{\sum_{c_k^+ \cup c_k^-}\exp( \langle \hat{f}(\hat{P}), \hat{f}(c) \rangle /\varphi)},
\label{equ:L_nce}
\end{equation}
where $c_k^+$ is the positive set and $c_k^-$ is the negative set for subset $c_k$. $\hat{f}(\cdot){=}g(\texttt{mp}(f(\cdot)))$ is a procedure consisting of the same backbone $f$ in the scorer network, a max-pooling operation $\texttt{mp}$ and a projection head $g$. It can project the pooled features of point subsets into the shared latent space.  
The overall loss used for training is simply given by:
\begin{equation}
    \mathcal{L}(\hat{P}) = \sum_{k=1}^{m} \mathcal{L}^{c_k}_{\mathrm{NCE}}(\hat{P}).
\end{equation} 
Due to the increasing cardinality of the subsets, the top points will be used more often in the contrastive loss, effectively scaling their importance for the total loss and leading to scores that put the most contrastively informative points at the top.

\section{Experimental setup}

\paragraph{Datasets.}
We evaluate our approach on three datasets.
\newline\textit{\textbf{ModelNet40}}~\cite{wu20153d} contains 40 categories from CAD models. We use the official split with 9,843 point clouds for training and 2,468 for testing. As the starting and maximum set, we uniformly sample 1,024 points from the mesh surface of each CAD model. 
\newline
\textit{\textbf{3D MNIST}}~\cite{rezende2016unsupervised} contains 6,000 raw point clouds, with each cloud containing 20,000 points. To enrich this dataset, we randomly select 1,024 points from each raw point cloud 10 times, resulting in a training set of size 50,000 and a test set of size 10,000, with each cloud consisting of 1,024 points. 
\newline
\textit{\textbf{ShapeNet Core55}}~\cite{chang2015shapenet} covers 55 object categories with 51,300 3D point clouds, with train/test sets following a 85\%/15\% split. We sample 1,024 points from the surface of each model to generate the initial set of data.

\begin{table}
	\centering
	\resizebox{1.0\columnwidth}{!}{%
		\begin{tabular}{llll}
			\toprule
			& \multicolumn{3}{c}{\textbf{Base Tasks}} \\
			\cmidrule(lr){2-4}
			& Classification & Retrieval & Reconstruction \\
			\midrule
			Training data & ModelNet40 & ModelNet40 & ModelNet40 \\
			Test data &  ModelNet40 & 3D MNIST & ShapeNet Core55 \\
			Base network & PointNet~\cite{qi2017pointnet} & PointNet++~\cite{qi2017pointnet++} & autoencoder~\cite{achlioptas2017learning} \\
			Metric & accuracy & mAP & reconstruction error \\
			\bottomrule
		\end{tabular}%
	}
	\caption{\textbf{Evaluation scheme}. We evaluate across three downstream tasks, three datasets, and three base networks.}
	\label{tab:evaluation}
\end{table}

\paragraph{Tasks and evaluation.}
Our method generates an ordering of points in terms of their importance in a self-supervised manner.
Given such an ordering, we consider three different downstream tasks for evaluation in an efficient manner: point cloud classification~\cite{qi2017pointnet}, retrieval~\cite{angelina2018pointnetvlad}, and reconstruction~\cite{achlioptas2017learning}. During training, we train our self-ordering approach on the ModelNet40 dataset without labels.
During evaluation, we firstly train the task networks on the corresponding test dataset as shown in Table~\ref{tab:evaluation} and keep the trained task networks fixed. Then we evaluate on variously sized subsets of points generated by the self-ordering model by passing them to the frozen task networks.
We report classification accuracy for classification, mean Average Precision (mAP) for retrieval, and the Chamfer distance~\cite{achlioptas2017learning} between the reconstructed points and original points for reconstruction, denoted as reconstruction error. Table~\ref{tab:evaluation} summarizes the evaluation tasks, test datasets, base networks for task and evaluation metrics.

\paragraph{Implementation details.}
For the Point Scorer network, we utilize a vanilla PointNet without any transformation layers (see Appendix for architectural details). 
We set the feature dimension size $D$ to 2,048, instead of the default 1,024. We vary the exact shape of the sigmoid function $\sigma(\cdot)$ by scaling the input with a temperature $\tau$ (Equation~\ref{equ:S_approx}). Unless otherwise noted, $\tau {=} 0.5$.
The Point Sorter is implemented using Sinkhorn-Knopp optimal transport, where  entropy regularisation parameter $\epsilon {=} 0.1$.
The backbones in the Hierarchical Contrastive Learning share weights with the backbone in the Point Scorer network for passing information and efficiency. 
A 2-layer MLP projection head projects all subset representations to a 128-dimensional latent space for self-supervised contrastive learning. 
We set the subset control factor $\theta{=}2$ and the temperature $\varphi$ in the NCE Loss function (Equation~\ref{equ:L_nce}) as $0{.}7$.

\paragraph{Training regime.} We train our self-ordering model, including the backbone, from scratch without any labels. We use a batch size 128 for 250 epochs unless otherwise noted, and use the AdamW optimizer~\cite{kingma2015adam} with weight decay  $1e{-}5$ and initialize the learning rate to $1e{-}4$ and decay the learning rate with the cosine decay schedule~\cite{loshchilov2016sgdr} to zero. Training on four Nvidia GTX 1080TI GPUs takes around 12 hours. Code is included in the supplementary materials.
\section{Results}
\subsection{Ablation study}

\begin{figure}[t!]
	\centering
	\includegraphics[width=0.99\columnwidth]{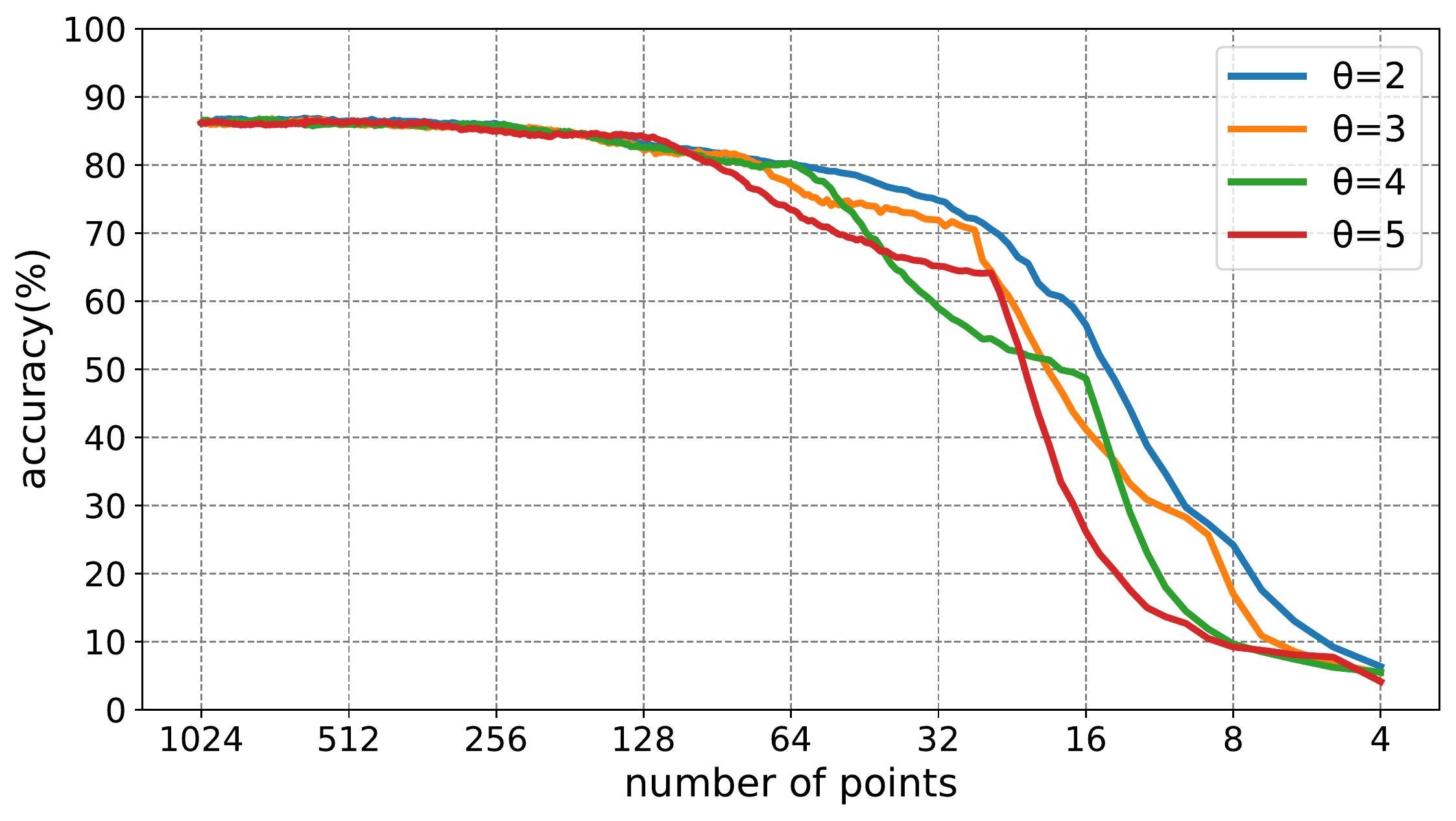}
	\caption{\textbf{Subset size controller $\theta$.} We evaluate for the classification task on ModelNet40. $\theta$ controls the size of the subsets for the Hierarchical Contrastive Learning. Having more subsets with a slower growth such as for 
	$\theta{=}2$ yields superior performances without kinks at powers of $\theta$.
	}
	\label{fig:delta}
\end{figure}

\paragraph{Subset size controller $\theta$.} The primary hyperparameter in our approach is the growth rate of the subsets on which the self-supervised contrastive loss are applied. We evaluate its influence from two (smallest growth) to five (largest growth). We show the classification results on ModelNet40 using PointNet as backbone in Figure~\ref{fig:delta}. We evaluate the classification effectiveness on point subsets across different sizes, ranging from all 1,024 points to the top 4 ranked points. We find that most of the performance is maintained with a cloud size from 1,024 to top 256 for all settings, indicating that only a subset of points is vital for classification. When retaining only the top 32 points, the accuracy is 73.5\% for $\theta{=}2$, compared to 65.7\%, 60.8\%, 67.8\% for respectively $\theta{=}3$, $\theta{=}4$, and $\theta{=}5$. The higher scores for the lowest $\theta$ value are a consequence of splitting the point cloud into more subsets in the Hierarchical Contrastive Learning. We conclude that using more subsets is beneficial for self-supervised ordering and use $\theta{=}2$ for all other experiments.

\begin{table}
	\centering
	\small 
		\begin{tabular}{lrrrr}
			\toprule
				&\multicolumn{4}{c}{\textbf{number of points}}\\
				\cmidrule(lr){2-5}
			$\tau$ & 16 & 32 & 64 & 128\\
			\midrule
	      	5 & 51.2 & 72.2 & 79.2 & 81.9\\
		    2 & 52.4 & 73.1 & 79.6 & 82.4\\
		    1 & 52.7 & 73.3 & 79.8 & 82.6\\
		    \rowcolor{mygray}
		  0.5 & 52.8 & 73.5 & 80.2 & {82.7}\\
		  0.1 & {52.9} & 73.4 & 79.8 & 82.3\\
			\bottomrule
		\end{tabular}%
	\caption{\textbf{Sigmoid temperature $\tau$}. We evaluate for the classification task on the ModelNet40 dataset. Results are stable for $\tau\le2$, for all number of points. An optimal result is obtained when $\tau$ is set to 0.5. }
	\label{tab:sigmoid}
\end{table}

\paragraph{Sigmoid temperature $\tau$.} Next we ablate the sensitivity of the differential approximation in the Point Scorer to the sigmoid temperature $\tau$. 
Table~\ref{tab:sigmoid} shows results of various $\tau$ values with evaluation for the classification task on ModelNet40.
%
%
We find that the performance of our method is quite stable with regards to this parameter even across orders of magnitude and find the best performance is achieved when the sigmoid is sharpened by a small amount with $\tau{=}0.5$, and use this value for all remaining experiments.
%

\begin{table}
	\centering
	\resizebox{0.86\columnwidth}{!}{%
		\begin{tabular}{rrrrrrr}
			\toprule
				&\multicolumn{4}{c}{\textbf{number of points}}& & \\
				\cmidrule(lr){2-5}
			$D$ & 16 & 32 & 64 & 128& CV & params(M)\\
			\midrule
		512 & 51.2 & 72.1 & 79.0 & 81.8 &2.1&1.2\\
		1,024 & 52.3 & 72.9 & 79.9 & 82.4 &1.7&1.7\\
		\rowcolor{mygray}
		2,048 & 52.8 & 73.5 & 80.2 & 82.7 &1.3&2.6\\
		4,096 & 52.9  & 73.8 & 80.4 & 82.8 &1.2&4.3\\
			\bottomrule
		\end{tabular}%
	}
	\caption{\textbf{Feature dimension size $D$.} We evaluate for the classification task on ModelNet40. Coefficient of Variance (CV) reflects the degree of dispersion among the point scores. Increasing dimension size $D$ avoids point scores to cluster at a value of 0 and improves the performances over all number of points, but at the expense of increased parameter cost. Balancing performance and parameters, we use $D{=}2,048$ for all experiments.}
	\label{tab:dimension}
\end{table}

\paragraph{Feature dimension size $D$.} The point scores come from the $D$-dimensional feature. If the dimension size $D$ is small, scores may degenerate to only a small subset of points with most points approaching the score of 0. A large $D$ helps avoid this degeneration, however, at the expense of an increased number of parameters. In Table~\ref{tab:dimension}, we ablate the influence of $D$ on the trade-of between performance and the amount of parameters. We evaluate for the classification task on ModelNet40 and include the Coefficient of Variance (CV), which reflects the degree of dispersion among the point scores. A high CV value means the points may aggregate around some lower score values.
When dimension size $D$ grows from 512 to 4,096, the CV value shrinks and performance increases. 
However, the CV gap between dimension of 2,048 and that of 4,098 is quite small,~\ie, 1.3~\textit{vs.} 1.2, and downstream performances are also close between the two dimension sizes,~\ie, 82.7~\textit{vs.} 82.8 when using the top 128 points. However, the parameters increase by 65\% from dimension size 2,048 to 4,096. Hence, 
we prefer to use 2,048 as the default value for the feature dimension size. 

\begin{table}
	\centering
\small
		\begin{tabular}{lrrrr}
			\toprule
				&\multicolumn{4}{c}{\textbf{number of points}}\\
				\cmidrule(lr){2-5}
			& 16 & 32 & 64 & 128\\
			\midrule
		scoring by max-pooling  & 36.3 & 51.6 & 66.2 & 73.7\\
		scoring by sum & 43.6 & 64.7 & 73.3 & 76.4\\
		\rowcolor{mygray}
		proposed Point Scorer & 52.8 & 73.5 & 80.2 & 82.7\\
			\bottomrule
		\end{tabular}%
	\caption{\textbf{Different point scoring methods}. We evaluate on classification for ModelNet40. Among three different  methods our proposed Point Scorer module produces the point scores leading to the best point cloud ordering.}
	\label{tab:scorer}
\end{table}

\paragraph{Different point scoring methods.} We tried three different methods to score the points: (i) we use a max-pooling operation on the feature map along the feature dimension, and obtain the max value of each point as point score; (ii) we use the sum of the feature value of each point as score; (iii) our proposed scorer  (Section~\textcolor{red}{3}). Results for classification on ModelNet40 in Table~\ref{tab:scorer} show our proposed scorer produces point scores with the best point cloud ordering.

\begin{table}
	\centering
\small
		\begin{tabular}{lcrrrr}
			\toprule
				& & \multicolumn{4}{c}{\textbf{number of points}}\\
				\cmidrule(lr){3-6}
			& sharing weights & 16 & 32 & 64 & 128\\
			\midrule
		\multirow{2}{*}{backbone} & no & 48.4 & 70.6 & 78.3 & 81.6\\
		 &\cellcolor{mygray} yes &\cellcolor{mygray} 52.8 &\cellcolor{mygray} 73.5 &\cellcolor{mygray} 80.2 &\cellcolor{mygray} 82.7\\
		\cmidrule(lr){1-6}
		\multirow{2}{*}{head} & no & 51.7 & 72.5 & 79.4 & 82.1\\
		 &\cellcolor{mygray} yes &\cellcolor{mygray} 52.8 &\cellcolor{mygray} 73.5 &\cellcolor{mygray} 80.2 &\cellcolor{mygray} 82.7\\
			\bottomrule
		\end{tabular}%
	\caption{\textbf{Sharing weights} in backbones and projection heads. We evaluate on classification and ModelNet40. We compare between sharing or not sharing weights for both backbones and projection heads. The result shows sharing weights both in backbones and in projection heads benefits our self-ordering module.}
	\label{tab:weights}
\end{table}

\paragraph{Sharing weights.} In Table~\ref{tab:weights} we compare between sharing and not sharing weights for the backbones (\ie for the Point Scorer and Hierarchical Contrastive Learning  and within the Hierarchical Contrastive Learning between different projection heads for the different subsets. 
We find that in both cases, sharing weights benefits point order learning, with the largest increase coming from a shared backbone between the scorer and the self-supervised loss module.
This is likely because similar features are required for both tasks and sharing weights encourages more general features to be favored and reinforced.

\begin{figure}[t!]
	\centering
	\includegraphics[width=0.99\columnwidth]{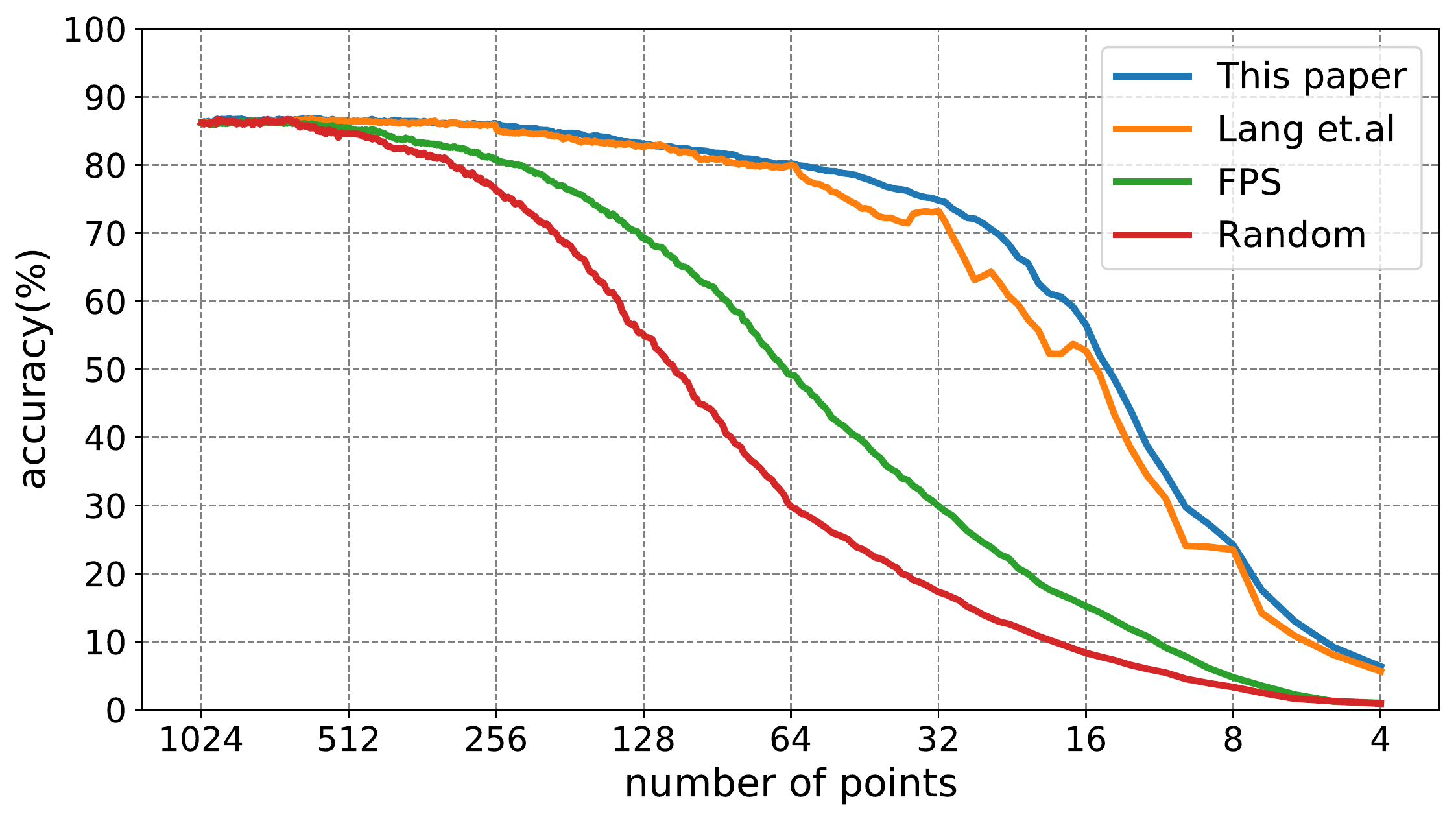}
	\caption{\textbf{Comparison on classification.} The instance classification accuracy is evaluated on consistent point numbers from all 1,024 points to the top 4 points. Our self-ordering method outperforms other ordering alternatives across all point numbers, including a fully-supervised alternative.}
	\label{fig:classification}
\end{figure}

\begin{table}[t]
	\centering
	\resizebox{0.99\columnwidth}{!}{%
		\begin{tabular}{lrrrrrrr}
			\toprule
				 &&\multicolumn{4}{c}{\textbf{Top \#pts from 1,024}}&time&memory\\
				\cmidrule(lr){3-6}
			& Training \#pts & 16 & 32 & 64 & 128&(hour)&(GB)\\
   \midrule
   random selection & - & 8.4 & 17.3 & 29.9 & 55.1 &-&-\\
		self-ordering & 256 & 50.4 & 71.5 & 78.9 & 81.3 &5.1&2.2\\
		self-ordering & 512 & 51.7 & 72.6 & 79.6 & 82.1 &9.7&3.7\\
		\rowcolor{mygray}
		self-ordering & 1,024 & 52.8 & 73.5 & 80.2 & 82.7 &12.3&6.1\\            
		\bottomrule
		\end{tabular}%
	}
	\caption{ \textbf{Scalability of ordering ability.}   We evaluate the classification task on ModelNet40 by varying the number of input points for our self-supervised training and selecting top points for evaluation from the 1,024-sized point cloud.
 }
	\label{tab:scalability}
\end{table}

\paragraph{Scalability of ordering ability.} We analyze the scalability of our approach in Table~\ref{tab:scalability}. We train self-ordering on point clouds of 256, 512 and 1,024 points and order point clouds containing 1,024 points in testing. The results remain stable, which demonstrates that self-ordering trained on small point clouds can generalize to order bigger point clouds. This enables ordering big point clouds with limited run time and GPU memory storage.


\subsection{Benchmarks}
We compare to three other methods able to obtain subsets of point clouds. First, a baseline based on random point selection~\cite{qi2017pointnet}; Second, a baseline based on Farthest Point Sampling (FPS)~\cite{li2018pointcnn,qi2019deep,qi2017pointnet++}, which starts from a random point in the cloud, and iteratively selects the farthest point from the selected points; Third, a fully supervised point sorter proposed by Lang~\etal~\cite{lang2020samplenet}. Note that different from Lang~\etal, we rely on self-supervision without the need for any cloud class labels or point-wise annotations prior to generating the ordering.

\paragraph{Classification.}
In Figure~\ref{fig:classification}, we show the comparison for classification on ModelNet40. We find our approach consistently outperforms the alternatives, independent of the number of points sampled for the downstream task. We even improve over the supervised point sorting by Lang \etal.

\begin{figure}[t!]
	\centering
	\includegraphics[width=0.99\columnwidth]{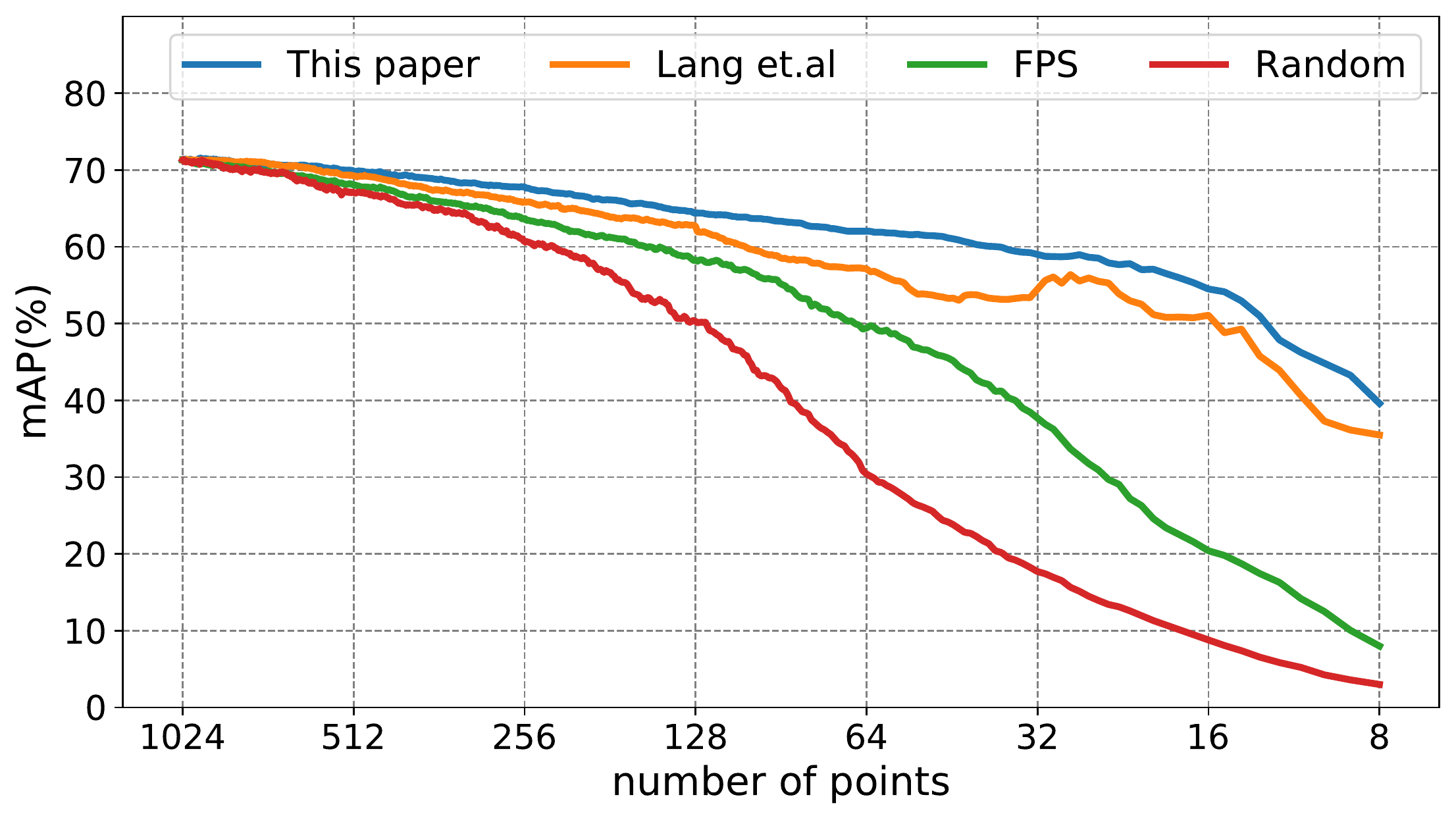}
	\caption{\textbf{Comparison on retrieval.} We measure retrieval mAP from all 1,024 points to the top 8 points based on the orderings. Our self-ordering performs better than the other three ordering methods independent of the number of points.}
	\label{fig:retrieval}
\end{figure}
\begin{figure}[!t]
	\centering
	\includegraphics[width=0.99\columnwidth]{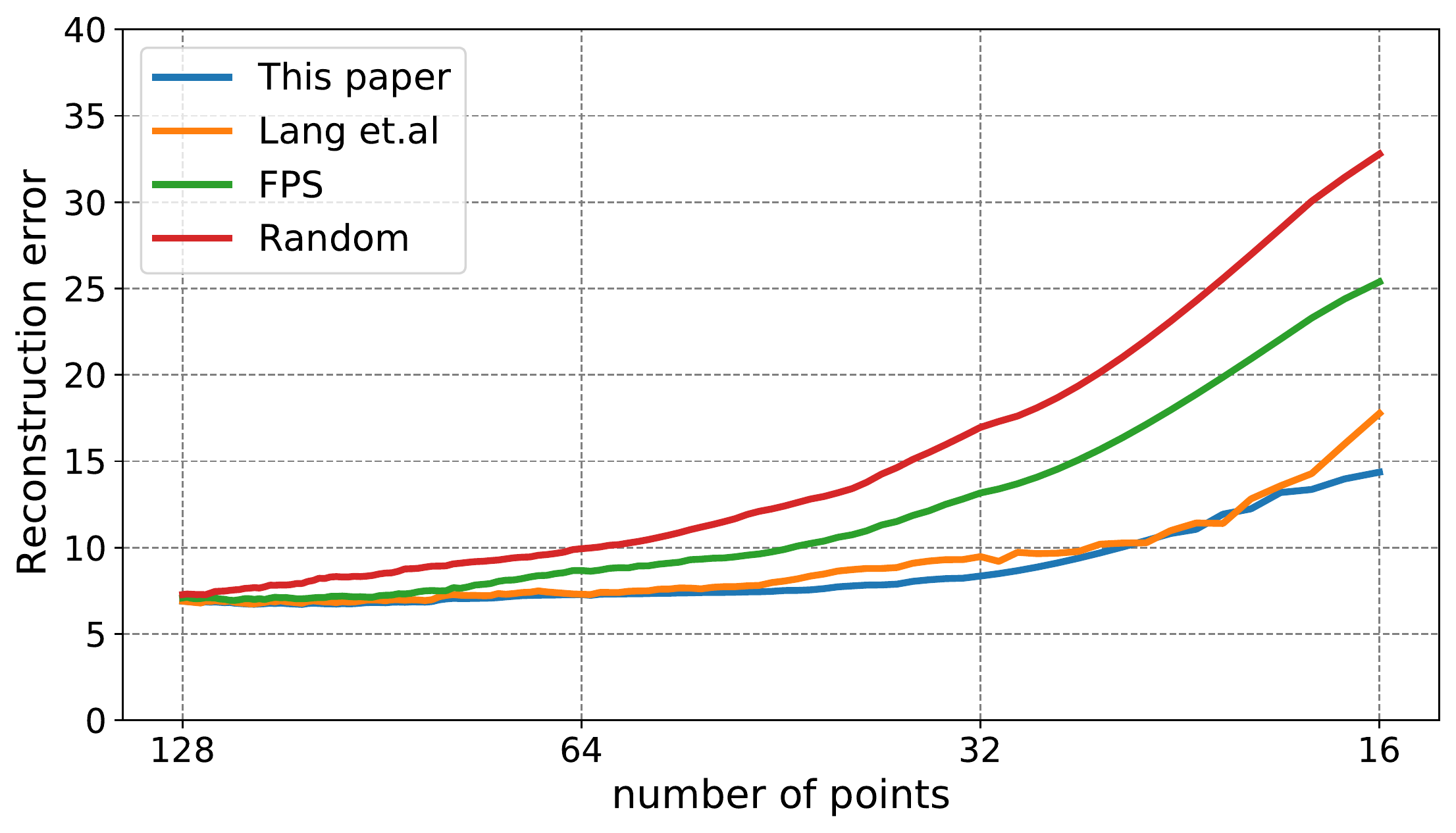}
	\caption{\textbf{Comparison on reconstruction.} We reconstruct from subsets of points selected by the orderings. From the top 128 points to the top 16 points, our self-ordering achieves a lower error compared to the alternatives.}
	\label{fig:reconstruction}
\end{figure}

\begin{figure*}[t]
	\centering
	\includegraphics[width=0.95\linewidth]{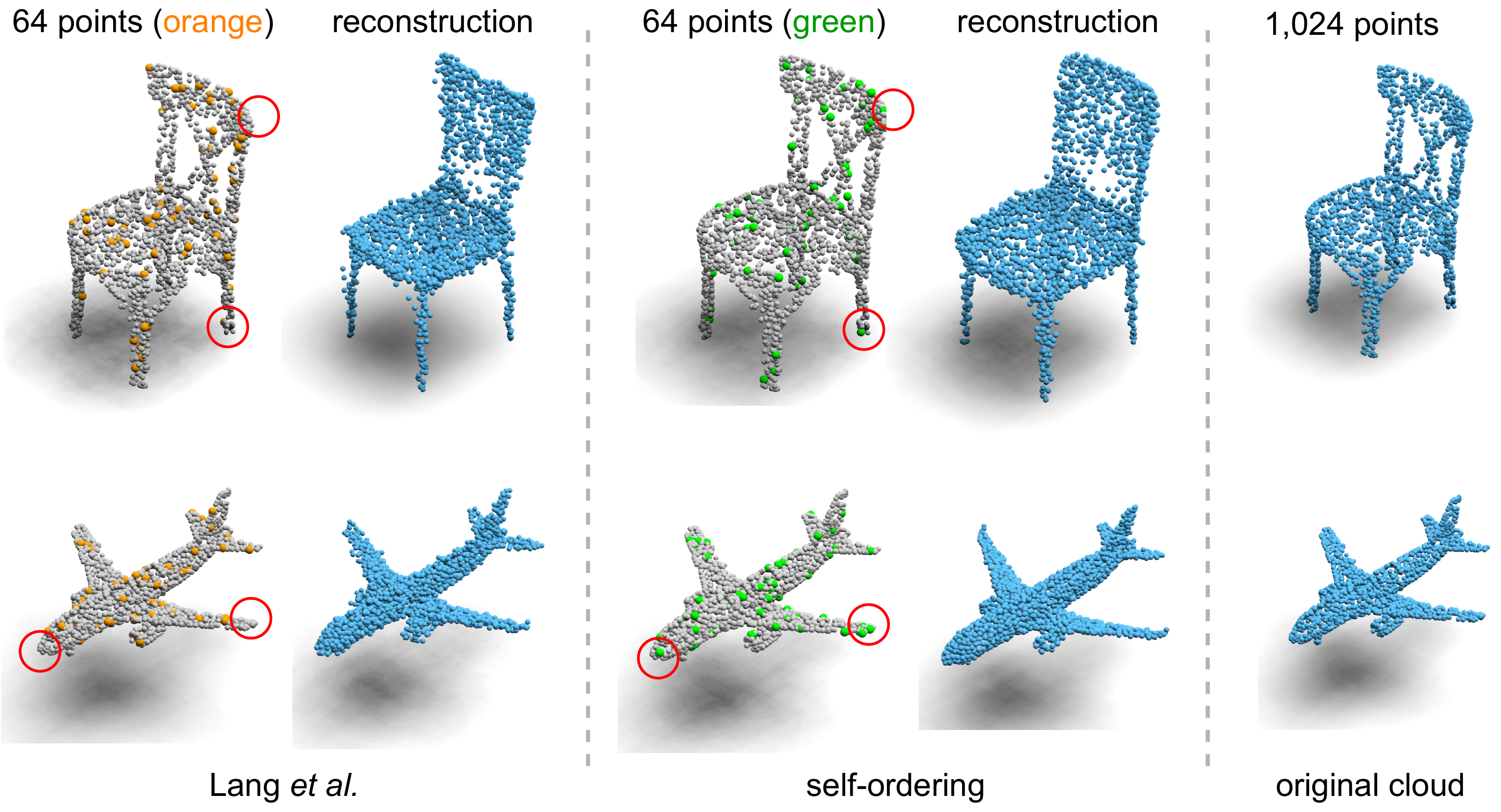}
	\caption{\textbf{Qualitative reconstruction results.} 
	We show our ability to recover the full, 1,024 points from only 64 points and the reconstruction from the supervised subset of Lang~\etal. Our top 64 points (\textcolor{green}{green points}) seem to pay more attention to the salient parts of objects compared with the top 64 points (\textcolor{orange}{orange points}) from the ordering of Lang~\etal (highlight in~\textcolor{red}{red circle}). This is reflected by the overall better reconstructions from our points when compared to the original. Further qualitative results are provided in the Appendix. }
	\label{fig:qualitative}
\end{figure*}

\paragraph{Retrieval.} Figure~\ref{fig:retrieval} presents the comparison for retrieval on 3D MNIST. We experiment on retrieval from all points to the top 8 points based on various orderings. We  observe considerably better performance when applying our self-ordering across all point numbers. For example, at top 8 points, we can achieve 39.6\% mAP, above the performance of 35.5\% obtained by the supervised method by Lang \etal.

\paragraph{Reconstruction.} In Figure~\ref{fig:reconstruction} we compare self-ordering with the three alternatives on the reconstruction task. A complete point cloud is reconstructed from the top 128 points up to the top 16 points based on the orderings. Our self-ordering achieve the lowest error across almost all point numbers, except for the top 24 and top 21 points, where the supervised method by Lang \etal overtakes self-ordering  (10.4 \textit{vs.} 10.3 and 11.9 \textit{vs.} 11.4). In Figure~\ref{fig:qualitative}, we show two examples of reconstructed point clouds from only 64 selected points, as provided by our self-ordering and the supervised ordering of Lang~\etal. We observe our top 64 points seem to pay more attention to the salient parts of objects compared with the top 64 points from the ordering of Lang~\etal. And, our reconstructed shapes are closer to the original shapes. The reconstructed chair of Lang~\etal becomes more square compared to the original round chair, while our reconstruction maintains curved edges. The reconstructed airplane of Lang~\etal has fuzzier edges compared with the original airplane cloud while our reconstructed airplane retains a sharp shape. 
The result further demonstrates the reconstruction ability of our method compared to the supervised alternative.

\begin{table}
	\centering
	\resizebox{0.99\columnwidth}{!}{%
\small 
       \setlength{\tabcolsep}{0.5em}
		\begin{tabular}{lcrrrr}
			\toprule
			& &\multicolumn{4}{c}{\textbf{number of points}}\\
			\cmidrule(lr){3-6}
			 & scenario & 16 & 32 & 64 & 128\\
			\midrule
			\rowcolor{mygray}
			\textbf{Classification} (\textit{accuracy}) & & & & &\\
			 \quad  random selection & - & 8.4 & 17.3 & 29.9 & 55.1\\
	      	\quad \multirow{2}{*}{self-ordering} & base & 52.8 & 73.5 & 80.2 & 82.7\\
                & zero-shot & 50.5 & 71.8 & 79.0 & 82.2\\
		    \rowcolor{mygray}
		    \textbf{Retrieval} (\textit{mAP}) & & & & &\\
		    \quad random selection & - & 8.8 & 17.7 & 30.4 & 50.3\\
		    \quad \multirow{2}{*}{self-ordering} & base & 54.5 & 59.0 & 62.2 & 64.7\\
                & zero-shot & 52.9 & 58.3 & 61.4 & 64.0\\
		    \rowcolor{mygray}
		    \textbf{Reconstruction} (\textit{error}) & & & & &\\
		    \quad random selection & - & 32.8 & 17.0 & 9.9 & 7.3\\
		    \quad \multirow{2}{*}{self-ordering}& base & 14.4 & 8.4 & 7.3 & 6.9\\ 
                & zero-shot & 15.1 & 8.9 & 7.7 & 7.0\\
			\bottomrule
		\end{tabular}%
	}
	\caption{\textbf{Zero-shot transfer.} We remove overlapping classes between training data and test data for the zero-shot scenario. The performance remains competitive when evaluating in this zero-shot point ordering scenario across three tasks and three datasets, demonstrating the generalization ability of our self-ordering. 
	} 
	\label{tab:overlap}
\end{table}

\paragraph{Zero-shot transfer.}
Finally, we evaluate the transfer performance of our self-ordering network. For this, we train on one dataset and evaluate the quality of the ordering of points it generates for a different dataset. We artificially create dataset splits that do not share common classes. To be precise, for ModelNet40 we generate a random 30/10-classes split. 
For the retrieval and reconstruction tasks, we remove the overlapping classes between the training and test datasets.
All class splits are listed in the Appendix. 
Experiments across three tasks and three datasets in Table~\ref{tab:overlap} show that the performances of the top points remain high even after removing the overlapping classes between the training and testing data. This shows that the self-ordering network has learned general features that can be used to transfer in a zero-shot fashion to unseen classes.

\section{Conclusion}

In this paper we tackled the problem of ordering points in a 3D point cloud such that their ranking can be used for selecting smaller subsets whilst retaining performance.
For this, we proposed a novel method that utilises  self-supervision to train a differentiable point orderer and evaluated its performance across three downstream tasks and datasets. The result is a principled, yet simple method that surpasses previous point selection heuristics by a large margin and even outperforms a supervised counterpart and enables ordering big point clouds in compute- and memory-constrained environments, showing the large potential that exists for self-supervised learning in point clouds.

{\small
\bibliographystyle{ieee_fullname}
\bibliography{7_egbib}
}

\end{document}